\documentclass[10pt,twocolumn,letterpaper]{article}

\usepackage{cvpr}              

\usepackage[dvipsnames]{xcolor}

\definecolor{cvprblue}{rgb}{0.21,0.49,0.74}
\usepackage[pagebackref,breaklinks,colorlinks,citecolor=cvprblue]{hyperref}
\usepackage{float}
\usepackage{placeins}

\def\paperID{*****} 
\def\confName{3DV\xspace}
\def\confYear{2026\xspace}

\title{XRefine: Attention-Guided Keypoint Match Refinement}

\author{  
	\thanks{Indicates equal contribution.}
	Jan Fabian Schmid\quad 
	\footnotemark[1] Annika Hagemann \\
	Bosch Research\\
	{\tt\small \{JanFabian.Schmid, Annika.Hagemann\}@de.bosch.com}
}

\begin{document}
\maketitle
\begin{abstract}
Sparse keypoint matching is crucial for 3D vision tasks, yet current keypoint detectors often produce spatially inaccurate matches. Existing refinement methods mitigate this issue through alignment of matched keypoint locations, but they are typically detector-specific, requiring retraining for each keypoint detector. We introduce \emph{XRefine}, a novel, detector-agnostic approach for sub-pixel keypoint refinement that operates solely on image patches centered at matched keypoints. Our cross-attention-based architecture learns to predict refined keypoint coordinates without relying on internal detector representations, enabling generalization across detectors.
Furthermore, XRefine can be extended to handle multi-view feature tracks. Experiments on MegaDepth, KITTI, and ScanNet demonstrate that the approach consistently improves geometric estimation accuracy, achieving superior performance compared to existing refinement methods while maintaining runtime efficiency. \\
Our code and trained models can be found at \url{https://github.com/boschresearch/xrefine}.
\end{abstract}    
\section{Introduction}
Extracting and matching sparse keypoints remain central to 3D computer vision systems, including structure-from-motion, visual localization, and SLAM. Despite the growing adoption of end-to-end, fully learned pipelines \cite{dust3r,mast3r,vggt}, many practical systems - particularly those with memory and runtime constraints - still depend on explicitly detected and matched keypoints. Sparse approaches offer clear benefits: they are lightweight, interpretable, and thus well-suited if dense inference is unnecessary or infeasible.\par

\begin{figure}
\centering
\includegraphics[width=0.46\textwidth]{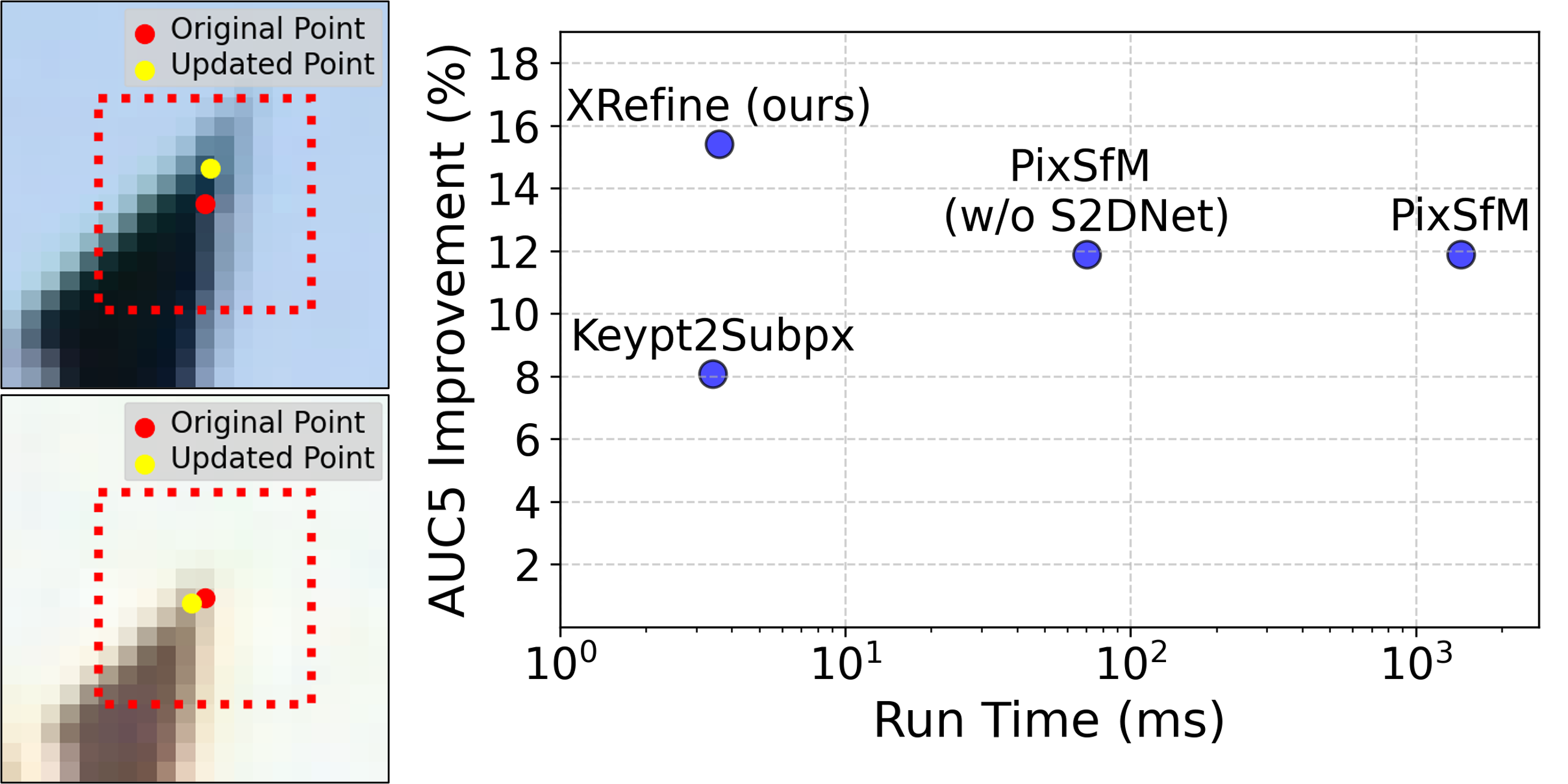}
\caption{\textbf{Attention-guided match refinement efficiently improves relative pose estimation}. \textbf{Left:} Exemplary matched SuperPoint~\cite{superpoint} keypoints on MegaDepth~\cite{megadepth}. The input to our model are the $11\times11$ patches within the red dotted lines. The refined keypoints of our model are presented as yellow dots. \textbf{Right:} Runtime and pose estimation improvement on MegaDepth (measured as relative increase in AUC5) of match refinement approaches averaged over five feature extractors: DeDoDe~\cite{dedode}, SIFT~\cite{sift}, SuperPoint~\cite{superpoint}, and XFeat~\cite{xfeat}. We compare our generalizing model to Keypt2Subpx~\cite{keypt2subpx} and the match refinement solution of PixSfM~\cite{pixsfm}. PixSfM extracts dense S2DNet~\cite{S2DNet} embeddings for feature-metric refinement. Depending on the use case this might be done exclusively for the refinement. Accordingly, we show the runtime of PixSfM with and without S2DNet inference.}
\label{fig:eye_catcher}
\end{figure}

\begin{figure*}[t]
\centering
\includegraphics[width=0.99\textwidth]{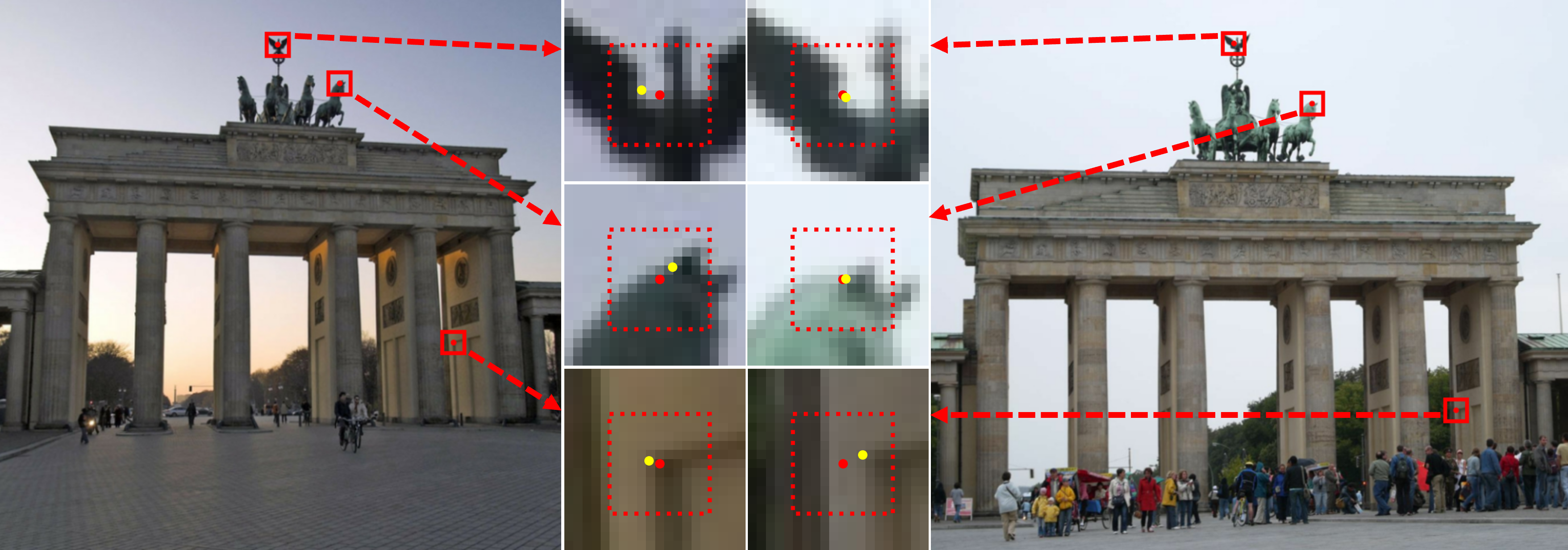}
\caption{Example match refinements from our model on MegaDepth~\cite{megadepth} for SuperPoint~\cite{superpoint} keypoints. The original keypoints are shown as red dots. In the magnified patches, the refined keypoints are shown as yellow dots. While the presented patches in this figure have a size of $21\times21$ pixels, the refinement model receives only the $11\times11$ area framed by the red dotted rectangle as input.}
\label{fig:intro_example_refinements}
\end{figure*}

The accuracy of keypoint-based systems is crucially influenced by the spatial accuracy of matched keypoints, \ie, how accurately the keypoints reflect the same physical 3D point geometrically (see \cref{fig:intro_example_refinements,fig:eye_catcher}). However, recent work \cite{keypt2subpx} shows that even state-of-the-art detectors suffer from inaccurate keypoint matches, decreasing geometric accuracy in downstream tasks. This limitation emerges naturally in keypoint detectors that only process each image separately, rendering the detection of keypoints at the exact same position in both images inherently difficult. \par

To address this limitation, recent refinement networks \cite{keypt2subpx, xfeat} adjust matched keypoint locations by simultaneously considering information of both images. Given a pair of matched keypoints, these models predict keypoint displacements using either keypoint descriptors \cite{xfeat} or scores and surrounding image patches \cite{keypt2subpx}. While these refinements improve the accuracy of downstream tasks like relative pose estimation, they require access to internal feature extractor representations (descriptors and keypoint scores). This necessity requires retraining for each detector architecture, limiting their generality and practical deployment.\par

We expand on this research by proposing a novel, detector-agnostic method for sub-pixel keypoint refinement called XRefine. Unlike the refinement networks in \cite{keypt2subpx, xfeat}, XRefine operates exclusively on image patches centered at matched keypoints \emph{without} requiring descriptors or keypoint scores. Thus, our model only needs to be trained once and generalizes across a wide range of classical (e.g., SIFT~\cite{sift}) and learned (e.g., SuperPoint~\cite{superpoint}, ALIKED~\cite{aliked}) detectors without requiring per-detector adaptation.\par

Inferring matched keypoint displacements solely from image patches requires information from both patches, which we realize using a cross-attention layer. Unlike existing image-patch-based refinement methods like PixSfM~\cite{pixsfm}, the proposed method does not rely on costly feature-metric optimization, but infers the refinement in a single forward pass. This makes XRefine lightweight and applicable on common edge AI chips.\par

Finally, we also propose a generalization of the approach from two-view matches to $n$-view feature tracks. This enables using the approach in SfM pipelines, as showcased for 3D point cloud triangulation on the ETH3D dataset~\cite{eth3d}.\par

We demonstrate that our approach consistently improves the accuracy of geometric estimation tasks across standard benchmarks such as MegaDepth~\cite{megadepth}, KITTI~\cite{kitti}, and ScanNet~\cite{scannet}, achieving higher pose accuracy than existing refinement approaches (see \cref{fig:eye_catcher}).\par 

In summary, our contributions are:
\begin{enumerate}
\item A cross-attention-based architecture for sub-pixel keypoint refinement that operates on image patches alone.
\item A detector-agnostic training scheme achieving generalization across a wide range of keypoint detectors.
\item A model variant for consistent multi-view refinement.
\item Superior performance across diverse datasets and feature extractors, without sacrificing runtime efficiency.
\end{enumerate}
\section{Related work}
\label{sec:related_work}

\paragraph{Sparse local feature extraction}
Tasks like camera pose estimation and calibration depend on the availability of point correspondences between images. Sparse local features are an efficient tool for determining correspondences:
\begin{enumerate}
    \item For each image, individually extract a set of keypoints along with corresponding score values and descriptors.
    \item Select the best keypoints per image based on their score.
    \item Identify potentially corresponding keypoints between images as those matched based on descriptor similarity.
\end{enumerate}

\begin{figure}
\centering
\includegraphics[width=0.46\textwidth]{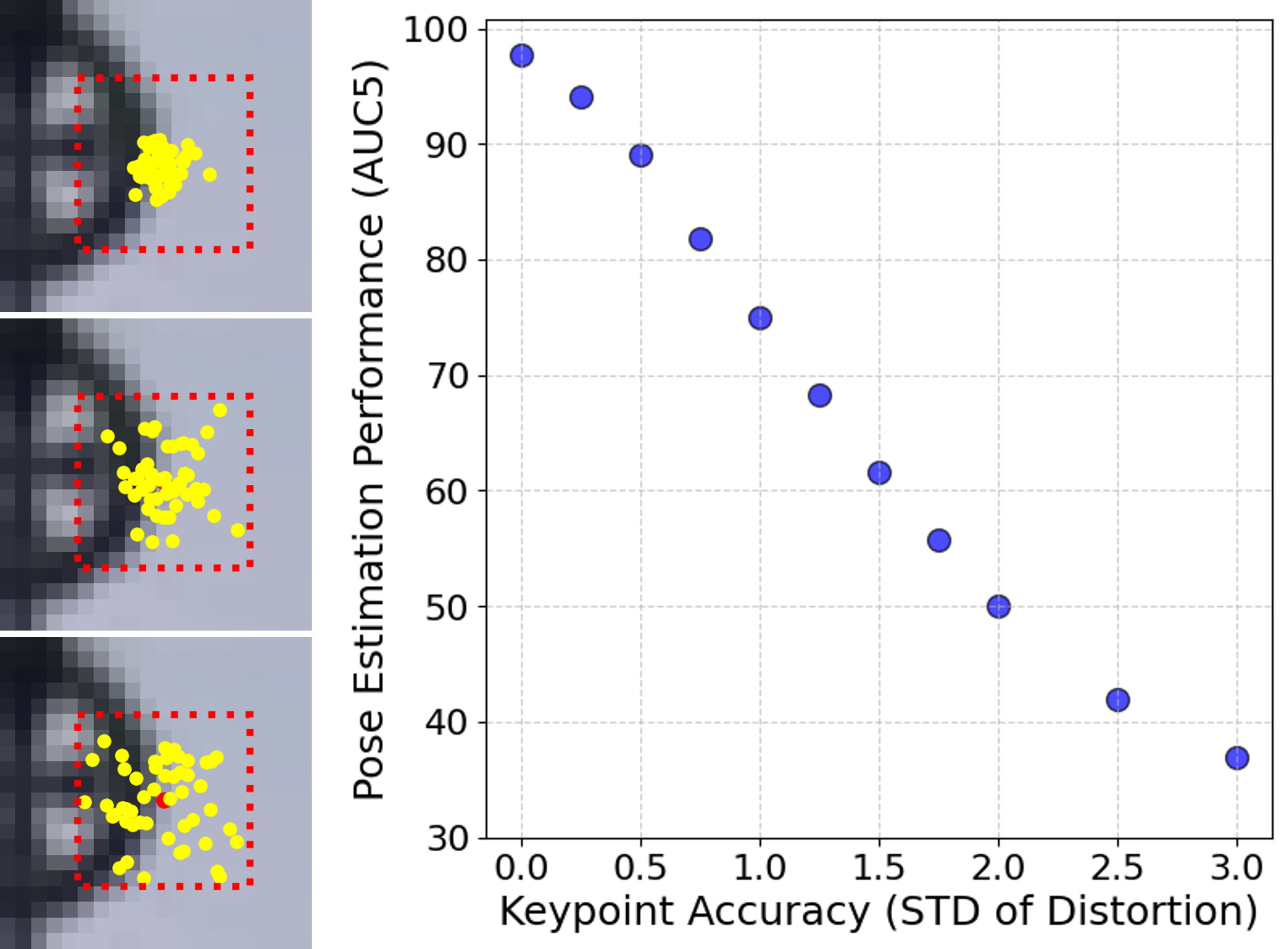}
\caption{Effect of inaccurate keypoint locations on the accuracy of relative pose estimation. \textbf{Left:} A patch of size $21\times21$ with a true keypoint shown as red dot and yellow dots representing sampled distortions to the keypoint (from top to bottom with a standard deviation of $1$, $2$, and $3$ pixels). The red dotted rectangle shows the $11\times11$ center area of the patch. \textbf{Right:} A graph illustrating the measured AUC5 pose estimation performance on the MegaDepth1500 dataset~\cite{megadepth}, using $2048$ ground truth correspondences perturbed with zero-mean Gaussian noise of varying standard deviation (STD) in pixels.}
\label{fig:auc_vs_std}
\end{figure}

Classical hand-crafted feature extraction, such as SIFT~\cite{sift}, detects keypoints as intensity extrema using a Difference of Gaussian pyramid. More recently, learning-based approaches started to outperform the classical approaches. DeTone \etal introduced SuperPoint~\cite{superpoint}, a fully-convolutional, single-forward-pass approach, leveraging Homographic Adaptation for self-supervised pre-training, which was later extended to fully self-supervised training in UnsuperPoint~\cite{unsuperpoint} and KP2D~\cite{kp2d}. Other methods focus on learning refined metrics, such as R2D2~\cite{r2d2}, which distinguishes descriptor reliability and keypoint repeatability, and DISK~\cite{disk}, which uses reinforcement learning to train the extractor end-to-end. Some feature extractors like DeDoDe~\cite{dedode} and DeDoDev2~\cite{dedodev2} aim for high performance, incorporating with DINOv2~\cite{dinov2} a large vision transformer as encoder. Efficiency-focused methods use lightweight CNN architectures, as in XFeat~\cite{xfeat}, or compute descriptors only at keypoint positions, like ALIKED~\cite{aliked}.
While the overall performance of local feature extraction has improved over time, recent work \cite{keypt2subpx} has shown that the spatial accuracy of keypoints still limits the accuracy of geometric downstream tasks (see~\cref{fig:auc_vs_std}).

\paragraph{Dense feature matching}
An alternative to the aforementioned sparse feature extraction methods are dense feature matching methods like LoFTR~\cite{loftr} and RoMa~\cite{roma}, which directly process image pairs. The availability of information from both images enables dense methods to outperform their sparse counter parts in terms of accuracy. However, dense matching approaches are computationally costly. Furthermore, extracting features independently in a first step can be advantageous, for example, in a Simultaneous Localization And Mapping (SLAM) context, where local features can be stored in a map to be matched with features of many other images recorded in the future. Match refinement techniques consider information from both images after matching and therefore have the potential to bridge the accuracy gap between sparse and dense approaches.

\paragraph{Approaches to match refinement}
Match refinement can be applied after feature matching to adjust the image coordinates of matched keypoints based on the assumption that they represent corresponding points. This is useful as even small inaccuracies of a single pixel or less can disturb resulting estimates, \eg of the camera pose (see~\cref{fig:auc_vs_std}). 

One class of approaches directly uses photometric alignment of local patches for match refinement, \eg Lucas–Kanade (LK) alignment \cite{lukas_kanade} and the inverse compositional LK \cite{ic_lk}. Such approaches, however, are computationally expensive and limited in their accuracy \cite{Lui_2018_CVPR}, particularly in cases of significant appearance changes.

Kim, Pollefeys, and Barath proposed Keypt2Subpx~\cite{keypt2subpx}, an efficient learning-based method for match refinement that leverages the corresponding image patches and descriptors of matched keypoints. The authors argue that their refinement method simplifies the keypoint detection task as it is no longer required to detect sub-pixel accurate keypoints. Subsequently, as it is done in SuperPoint~\cite{superpoint} and XFeat~\cite{xfeat}, the extractor can save computational effort by providing pixel coordinates as keypoints. Keypt2Subpx is trained to minimize the epipolar error. Accordingly, instead of requiring ground truth coordinates for matched keypoints, it is sufficient to have ground truth essential matrices for given image pairs, allowing the model to optimize keypoint positions directly for camera pose estimation.

Dusmanu \etal~\cite{local_feature_refinement} propose Patch Flow, a refinement approach that aligns patches based on local optical flow and its resulting geometric cost.
Lindenberger, Sarlin \etal improve upon Patch Flow with PixSfM~\cite{pixsfm}, which presents a solution for match refinement in a multi-view scenario. They identify matches of the same keypoint over multiple images as tracks and then adjust the coordinates of all involved keypoints jointly in a featuremetric optimization.

As described previously, the feature extractor XFeat~\cite{xfeat} detects keypoints only at pixel accuracy. However, Potje \etal propose a learned match refinement module that takes only the descriptors of matched keypoints as input and provides a sub-pixel offset as output that is added to the keypoints to improve their accuracy.

\begin{figure*}[ht]
\centering
\includegraphics[width=0.99\textwidth]{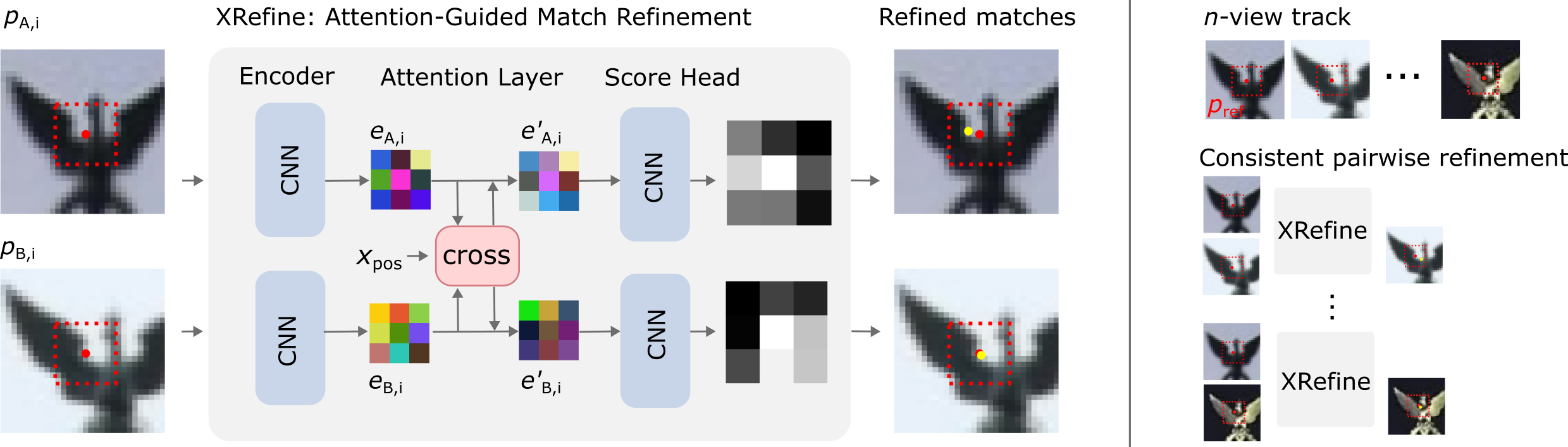}
\caption{Architecture of our attention-guided match refinement. \textbf{Left:} The model takes $11\times11$ image patches $p_{A,i}, p_{B,i}$ (red dotted rectangle) around matched keypoints (red dots) as input. A CNN extracts embeddings $e_{A,i}, e_{B,i}$ which are updated using cross-attention. The score head then maps the updated embeddings $e'_{A,i}, e'_{B,i}$ to score maps $S_{A,i}, S_{B,i}$. A soft-argmax operation on these score maps finally yields the updated keypoint positions (yellow dots). \textbf{Right:} Extension to $n$-view problems. By using one patch as reference $p_{\mathrm{ref}}$ and using a model variant that refines only the second (non-reference) keypoint, consistent refinements can be obtained.}
\label{fig:architecture}
\end{figure*}

The match refinement solution presented in this paper differs from these approaches in several aspects. In contrast to Keypt2Subpx~\cite{keypt2subpx} and XFeat~\cite{xfeat}, our model takes only image patches at keypoint positions as input and not the descriptors or other output of the feature extractor, like the keypoint score. Hence, our model does not have to be trained specifically for each feature extractor. 
Unlike PixSfM~\cite{pixsfm}, our method does not rely on costly feature-metric optimization, but infers the refinement, using a light-weight neural network, in a single forward pass. This makes the approach fast, while giving highest accuracy in match refinement across feature extractors (\cref{fig:eye_catcher}).

\section{Method}
\label{sec:method}

We present \emph{XRefine}, an attention-based keypoint match refinement model that takes only image patches as input and provides adjusted keypoint positions as output. For best generalizability, the model is trained feature extractor independently; we refer to this variant as \textbf{XRefine general}. For best accuracy, the model can be trained specifically for a feature extractor; we refer to this variant as \textbf{XRefine specific}. An overview of the approach is presented in \cref{fig:architecture}.

\subsection{Architecture}
The model takes two gray-scale patches $p_A$ and $p_B$ of size $11\times11$ as input. Both patches are processed independently by an encoder. The encoder performs five convolutions with $3\times3$ kernels, increasing the channel size from $1$ to $16$ with the first operation and then to $64$ with the third operation. The first three and the last convolution are executed without padding; hence, the final embeddings $e_A$ and $e_B$ have a size of only $3\times3$. 
Now, a single block of multi-head cross-attention is applied between the two patch embeddings $e_A$ and $e_B$. Each embedding is translated into a sequence of $3\times3=9$ tokens of dimensionality $64$. To provide spatial information for each token, a learned positional encoding $x_{\mathrm{pos}}$ is added to the sequences. In order to update $e_A$, we use $e_A$ as query and $e_B$ as key and value, and vise versa to update $e_B$.
After the cross attention, a score map head individually takes the updated embeddings as input, outputting their respective score map. The score map head is performing a single convolution with kernel size $3\times3$, with padding to keep the same size for the output. Then, a tanh operation brings the values into a range of $[-1, 1]$.   
Finally, similarly as in \cite{keypt2subpx}, a spatial soft-argmax is applied to each score map to obtain the updated keypoint position. The resulting coordinates are interpreted as relative coordinates to the center of the original patch. Accordingly, they are scaled-up to represent positions in the original $11\times11$ patch.   

\subsection{Training}
The model is trained with the geometric training objective proposed by Kim, Pollefeys, and Barath~\cite{keypt2subpx}, optimizing the epipolar error directly.

\paragraph{Dataset generation}
\label{sec:method:dataset_generation}
Training our models requires image pairs with overlapping field-of-view and known relative poses. We use two different training paradigms for the \emph{specific} and \emph{general} model:
For the feature extractor specific datasets for \emph{XRefine specific}, we use the respective feature extractor and detect $4096$ keypoints with highest score values in each image. They are then matched, using mutual nearest neighbor matching (MNN), double soft max (DSM)~\cite{dedode}, or LightGlue~\cite{lightglue}, depending on the extractor. To train \emph{XRefine general}, we randomly select $4096$ pixel coordinates with available depth information in the first image and project it into the second image to create a matching pair of \textit{keypoints}. Then, both keypoints are randomly perturbed by adding a vector with x and y values sampled from a zero-mean normal distribution with a standard deviation of $1.5$ pixels. We also tested smaller and larger standard deviations, as well as a uniform distribution, but observed best results with this setting. Subsequently, $11\times11$ image patches are cropped at the center of each matched keypoint. \par

We train on MegaDepth~\cite{megadepth}, splitting the dataset as in \cite{loftr} into $45900$ train samples, $655778$ evaluation samples, and $1500$ samples for validation (also referred to as MegaDepth1500). Each sample represents two views with partially overlapping content. Images are loaded with the GlueFactory~\cite{gluestick} library, resizing them to $1024$ pixels on the longer side, while keeping the aspect ratio.

\paragraph{Details}
\label{sec:method:details}
Our training runs for $120$ epochs. In each epoch, $2048$ matches are randomly sampled for each image pair of the training split of MegaDepth~\cite{megadepth}. We use PyTorch 2.1.2, the Adam optimizer~\cite{adam} with a learning rate of $0.0001$, and a batch size of $8$. We validate after each epoch on MegaDepth1500~\cite{megadepth}. The weights for a given setup are selected as those with highest AUC5 performance on the validation dataset within two trainings with different seeds.

\subsection{Generalization to $n$ images}
\label{sec:method:n_images}
The proposed model adjusts keypoint locations across two views. However, some 3D vision tasks require consistent keypoints across $n$ views. One example is Structure-from-Motion which typically builds feature tracks consisting of $T\geq 2, T\in \mathbb{N}$ matched keypoints. Naively applying our refinement to individual image pairs within a track could result in inconsistent refinements across pairs. To address this issue, we propose an architecture variant which only adjusts the second keypoint (see \cref{fig:architecture}). In this variant, we still perform cross-attention between feature maps, but the score map $S_i$ as well as the keypoint shift $d_i$ are only inferred for the second image.\par

Given a feature track $\{\mathbf{u}_1, \mathbf{u}_2, ... \mathbf{u}_T\}$, we then define one of the keypoints as reference $\mathbf{u}_{\mathrm{ref}}$, and apply the refinement to all other keypoints by passing pairs $\{(\mathbf{u}_{\mathrm{ref}}, \mathbf{u}_2), (\mathbf{u}_{\mathrm{ref}}, \mathbf{u}_3),... (\mathbf{u}_{\mathrm{ref}}, \mathbf{u}_{T-1})\}$ to the model. Thereby, all keypoints are refined towards the reference keypoint, resulting in a consistently refined track.
\section{Evaluation}

\label{sec:evaluation}
We evaluate match refinement for relative pose estimation in \cref{sec:evaluation:pose_estimation} and point cloud triangulation in \cref{sec:evaluation:triangulation}.

\begin{table}
\centering
\begin{tabular}{@{}l@{\hspace{4pt}}c@{\hspace{4pt}}c@{\hspace{4pt}}c@{\hspace{4pt}}c@{}}
\toprule
Dataset & Refinement & Avg.(\%) & Min.(\%) & Max.(\%) \\
\midrule
MegaDepth & Keypt2Subpx & 8.07 & 0.91 & 28.61 \\
MegaDepth & PixSfM & 11.92 & 0.30 & 33.60 \\
MegaDepth & XRefine general & 15.42 & 2.28 & 42.26 \\
MegaDepth & XRefine specific & 15.99 & 3.80 & 41.92 \\
\hline
ScanNet & Keypt2Subpx & 6.26 & -0.51 & 10.56 \\
ScanNet & PixSfM & 8.74 & -3.77 & 17.81 \\
ScanNet & XRefine general & 16.34 & 3.43 & 27.11 \\
ScanNet & XRefine specific & 17.52 & 4.29 & 29.11 \\
\hline
KITTI & Keypt2Subpx & 0.23 & -0.43 & 1.07 \\
KITTI & PixSfM & 0.75 & -0.72 & 2.01 \\
KITTI & XRefine general & 1.07 & -0.38 & 2.91 \\
KITTI & XRefine specific & 1.18 & -0.21 & 2.73 \\
\bottomrule
\end{tabular}
\caption{
Summary of results for the extractors DeDoDe~\cite{dedode}, SIFT~\cite{sift}, SuperPoint~\cite{superpoint}, and XFeat~\cite{xfeat} on each of our three evaluation datasets MegaDepth~\cite{megadepth}, ScanNet~\cite{scannet}, and KITTI~\cite{kitti}. We present the average, minimum, and maximum improvement of the AUC5 relative to the results without match refinement. 
}
\label{tab:evaluation:averages}
\end{table}

\subsection{Relative pose estimation}
\label{sec:evaluation:pose_estimation}
We evaluate on the photo-tourism dataset MegaDepth~\cite{megadepth}, the indoor dataset ScanNet~\cite{scannet}, and the KITTI~\cite{kitti} visual odometry dataset. Our use of MegaDepth~\cite{megadepth} is described in \cref{sec:method:dataset_generation}. Due to the large size of the evaluation dataset, we consider only every 10th image pair, \ie $65577$ pairs. For ScanNet~\cite{scannet}, we evaluate on the $1500$ samples selected in \cite{superglue}, resizing images to $640\times480$. For KITTI, we use the $2790$ image pairs selected in \cite{9341229} at a size of $1240\times376$.

We compare \emph{XFeat specific} and \emph{XFeat general} with three state-of-the-art match refinement approaches described in \cref{sec:related_work}: Keypt2Subpx~\cite{keypt2subpx}, PixSfM~\cite{pixsfm}, and the refinement approach proposed in XFeat~\cite{xfeat}. For Keypt2Subpx, weights for only a few feature extractors are publicly available; therefore, we train the model with the same procedure described in \cref{sec:method:details}. We observe very similar performance of our re-trained Keypt2Subpx weights as for the publicly available weights. Details can be found in the appendix. The PixSfM solution for match refinement is independent of the feature extractor, so we can use the publicly available solution. The XFeat refinement approach is trained specifically for a variant of XFeat~\cite{xfeat} that is called XFeat*, which, in contrast to the default XFeat, extracts features at two image sizes, and is reported to achieve better performance when using a larger number of features per image. We use the weights provided by the authors and use the XFeat solution only for XFeat*.

If not specified differently, we extract always $2048$ features per image and match features using mutual nearest neighbor matching (MNN), double soft max (DSM)~\cite{dedode}, or LightGlue (LG)~\cite{lightglue}. For essential matrix estimation, we employ, as suggested in \cite{keypt2subpx}, GC-RANSAC~\cite{GC_RANSAC} with $1000$ iterations and a threshold of $1$~pixel. 

In terms of evaluation metrics, we follow \cite{jin2021image}, measuring pose estimation performance as area under curve (AUC) of pose errors that represent the maximum of translation direction error and the rotation error of the estimated pose compared to the given ground truth. We report the AUC for thresholds of $5$, $10$, and $20$ degrees. The reported values are averages from $10$ repetitions of the same experiment.

\begin{table}
\centering
\begin{tabular}{@{}l@{\hspace{4pt}}c@{\hspace{4pt}}c@{\hspace{4pt}}c@{\hspace{4pt}}c@{}}
\toprule
Extract+Match & Refinement & AUC5 & AUC10 & AUC20 \\
\midrule
SP+MNN &  & 34.91 & 45.00 & 53.56 \\
SP+MNN & Keypt2Subpx & 36.20 & 46.09 & 54.32 \\
SP+MNN & PixSfM & 38.30 & 47.96 & 55.86 \\
SP+MNN & XRefine general & \textbf{38.87} & \textbf{48.50} & \underline{56.35} \\
SP+MNN & XRefine specific & \underline{38.86} & \textbf{48.50} & \textbf{56.36} \\
\hline
SP+LG &  & 58.48 & 71.41 & 80.83 \\
SP+LG & Keypt2Subpx & 60.16 & 72.73 & 81.78 \\
SP+LG & PixSfM & 62.05 & 74.15 & 82.72 \\
SP+LG & XRefine general & \underline{62.86} & \underline{74.83} & \underline{83.27} \\
SP+LG & XRefine specific & \textbf{63.07} & \textbf{75.02} & \textbf{83.40} \\
\hline
DeDoDe+DSM &  & 34.88 & 48.64 & 60.84 \\
DeDoDe+DSM & Keypt2Subpx & 44.86 & 58.08 & 68.84 \\
DeDoDe+DSM & PixSfM & 46.60 & 59.09 & 69.20 \\
DeDoDe+DSM & XRefine general & \textbf{49.62} & \textbf{62.20} & \underline{72.06} \\
DeDoDe+DSM & XRefine specific & \underline{49.50} & \underline{62.12} & \textbf{72.07} \\
\hline
SIFT+MNN &  & 19.76 & 26.53 & 33.00 \\
SIFT+MNN & Keypt2Subpx & 19.94 & 26.67 & 33.09 \\
SIFT+MNN & PixSfM & 19.82 & 26.49 & 32.82 \\
SIFT+MNN & XRefine general & \underline{20.21} & \underline{26.95} & \underline{33.31} \\
SIFT+MNN & XRefine specific & \textbf{20.51} & \textbf{27.31} & \textbf{33.69} \\
\hline
XFeat+MNN &  & 36.45 & 47.89 & 57.81 \\
XFeat+MNN & Keypt2Subpx & 38.01 & 49.06 & 58.52 \\
XFeat+MNN & PixSfM & 40.06 & 50.99 & 60.13 \\
XFeat+MNN & XRefine general & \underline{41.46} & \underline{52.37} & \underline{61.35} \\
XFeat+MNN & XRefine specific & \textbf{41.94} & \textbf{52.83} & \textbf{61.78} \\
\hline
XFeat*+MNN &  & 18.80 & 30.26 & 42.25 \\
XFeat*+MNN & XFeat-Refine. & 31.18 & 43.25 & 54.15 \\
XFeat*+MNN & Keypt2Subpx & 33.11 & 45.18 & 55.96 \\
XFeat*+MNN & PixSfM & 34.63 & 46.07 & 56.16 \\
XFeat*+MNN & XRefine general & \underline{37.96} & \underline{49.62} & \underline{59.60} \\
XFeat*+MNN & XRefine specific & \textbf{38.36} & \textbf{49.99} & \textbf{59.90} \\
\bottomrule
\end{tabular}
\caption{Pose estimation results on MegaDepth~\cite{megadepth}. Bold indicates best performance and underscores second best per feature.}
\label{tab:evaluation:mega}
\end{table}

\begin{table}
\centering
\begin{tabular}{@{}l@{\hspace{4pt}}c@{\hspace{4pt}}c@{\hspace{4pt}}c@{\hspace{4pt}}c@{}}
\toprule
Extract+Match & Refinement & AUC5 & AUC10 & AUC20 \\
\midrule
SP+MNN &  & 11.51 & 23.87 & 37.92 \\
SP+MNN & Keypt2Subpx & 12.36 & 25.07 & 39.20 \\
SP+MNN & PixSfM & 13.56 & 26.78 & 40.66 \\
SP+MNN & XRefine general & \underline{14.63} & \underline{28.23} & \underline{42.33} \\
SP+MNN & XRefine specific & \textbf{14.86} & \textbf{28.38} & \textbf{42.48} \\
\hline
SP+LG &  & 19.48 & 37.40 & 54.79 \\
SP+LG & Keypt2Subpx & 20.31 & 38.21 & 55.43 \\
SP+LG & PixSfM & 21.25 & 39.29 & 55.93 \\
SP+LG & XRefine general & \textbf{22.49} & \textbf{40.58} & \textbf{57.27} \\
SP+LG & XRefine specific & \underline{21.90} & \underline{40.10} & \underline{56.83} \\
\hline
DeDoDe+DSM &  & 10.13 & 21.04 & 32.42 \\
DeDoDe+DSM & Keypt2Subpx & 11.20 & 23.02 & 34.99 \\
DeDoDe+DSM & PixSfM & 10.44 & 21.85 & 33.98 \\
DeDoDe+DSM & XRefine general & \underline{11.55} & \underline{23.15} & \underline{35.27} \\
DeDoDe+DSM & XRefine specific & \textbf{11.71} & \textbf{23.55} & \textbf{35.51} \\
\hline
SIFT+MNN &  & 5.83 & 12.32 & 20.14 \\
SIFT+MNN & Keypt2Subpx & 5.80 & 12.37 & 20.16 \\
SIFT+MNN & PixSfM & 5.61 & 11.88 & 19.36 \\
SIFT+MNN & XRefine general & \underline{6.03} & \underline{12.71} & \underline{20.65} \\
SIFT+MNN & XRefine specific & \textbf{6.08} & \textbf{12.79} & \textbf{20.72} \\
\hline
XFeat+MNN &  & 10.28 & 22.04 & 35.77 \\
XFeat+MNN & Keypt2Subpx & 11.27 & 23.32 & 37.08 \\
XFeat+MNN & PixSfM & 12.08 & 24.40 & 38.23 \\
XFeat+MNN & XRefine general & \underline{12.51} & \underline{25.49} & \underline{39.43} \\
XFeat+MNN & XRefine specific & \textbf{12.97} & \textbf{26.07} & \textbf{40.02} \\
\hline
XFeat*+MNN &  & 8.32 & 18.65 & 32.21 \\
XFeat*+MNN & XFeat-Refine. & 12.89 & 26.30 & 41.17 \\
XFeat*+MNN & Keypt2Subpx & 13.23 & 26.29 & 40.66 \\
XFeat*+MNN & PixSfM & 12.42 & 24.99 & 39.25 \\
XFeat*+MNN & XRefine general & \textbf{14.16} & \underline{27.23} & \underline{41.63} \\
XFeat*+MNN & XRefine specific & \textbf{14.16} & \textbf{27.50} & \textbf{41.99} \\
\bottomrule
\end{tabular}
\caption{Pose estimation results on ScanNet~\cite{scannet}. Bold indicates best performance and underscores second best per feature.}
\label{tab:evaluation:scannet}
\end{table}

\begin{table}
\centering
\begin{tabular}{@{}l@{\hspace{4pt}}c@{\hspace{4pt}}c@{\hspace{4pt}}c@{\hspace{4pt}}c@{}}
\toprule
Extract+Match & Refinement & AUC5 & AUC10 & AUC20 \\
\midrule
SP+MNN &  & 83.11 & 90.74 & 95.16 \\
SP+MNN & Keypt2Subpx & 83.07 & 90.70 & 95.15 \\
SP+MNN & PixSfM & 84.11 & 91.28 & 95.44 \\
SP+MNN & XRefine general & \underline{84.22} & \underline{91.33} & \underline{95.46} \\
SP+MNN & XRefine specific & \textbf{84.37} & \textbf{91.40} & \textbf{95.49} \\
\hline
SP+LG &  & 83.37 & 90.84 & 95.12 \\
SP+LG & Keypt2Subpx & 83.63 & 90.93 & 95.14 \\
SP+LG & PixSfM & 84.22 & 91.30 & 95.39 \\
SP+LG & XRefine general & \underline{84.34} & \underline{91.35} & \textbf{95.40} \\
SP+LG & XRefine specific & \textbf{84.42} & \textbf{91.38} & \textbf{95.40} \\
\hline
DeDoDe+DSM &  & 83.98 & 91.31 & 95.42 \\
DeDoDe+DSM & Keypt2Subpx & 84.21 & \underline{91.42} & \underline{95.50} \\
DeDoDe+DSM & PixSfM & 84.17 & 91.35 & 95.45 \\
DeDoDe+DSM & XRefine general & \underline{84.24} & \underline{91.42} & \underline{95.50} \\
DeDoDe+DSM & XRefine specific & \textbf{84.50} & \textbf{91.54} & \textbf{95.57} \\
\hline
SIFT+MNN &  & \textbf{83.79} & \textbf{91.32} & \textbf{95.51} \\
SIFT+MNN & Keypt2Subpx & 83.43 & 91.14 & 95.44 \\
SIFT+MNN & PixSfM & 83.19 & 91.03 & 95.39 \\
SIFT+MNN & XRefine general & 83.47 & 91.17 & 95.45 \\
SIFT+MNN & XRefine specific & \underline{83.61} & \underline{91.24} & \underline{95.49} \\
\hline
XFeat+MNN &  & 81.55 & 89.99 & 94.79 \\
XFeat+MNN & Keypt2Subpx & 82.42 & 90.47 & 95.00 \\
XFeat+MNN & PixSfM & 83.19 & 90.93 & 95.27 \\
XFeat+MNN & XRefine general & \textbf{83.92} & \textbf{91.26} & \textbf{95.39} \\
XFeat+MNN & XRefine specific & \underline{83.78} & \underline{91.21} & \underline{95.37} \\
\hline
XFeat*+MNN &  & 13.79 & 19.99 & 24.07 \\
XFeat*+MNN & XFeat-Refine. & 77.59 & 86.98 & 92.55 \\
XFeat*+MNN & Keypt2Subpx & 72.99 & 84.01 & 90.43 \\
XFeat*+MNN & PixSfM & 78.73 & 87.90 & 93.17 \\
XFeat*+MNN & XRefine general & \underline{80.63} & \underline{89.05} & \underline{93.85} \\
XFeat*+MNN & XRefine specific & \textbf{80.90} & \textbf{89.30} & \textbf{94.11} \\
\bottomrule
\end{tabular}
\caption{Pose estimation results on KITTI~\cite{kitti} odometry. Bold indicates best performance and underscores second best per feature.}
\label{tab:evaluation:kitti}
\end{table}

\paragraph{Main results} 
As a brief overview, \cref{tab:evaluation:averages} summarizes the results over the evaluated feature extractor and matcher pairings, including DeDoDe~\cite{dedode}, SIFT~\cite{sift}, SuperPoint (SP)~\cite{superpoint}, and XFeat~\cite{xfeat}. 
Individual results are shown in \cref{tab:evaluation:mega} for MegaDepth~\cite{megadepth}, \cref{tab:evaluation:scannet} for ScanNet~\cite{scannet}, and \cref{tab:evaluation:kitti} for KITTI~\cite{kitti}. More feature extractors are presented in the appendix. Results for XFeat* are not included in the summary \cref{tab:evaluation:averages}, as it is an outlier extractor that was not intended to be used without match refinement and therefore has an unusually large benefit from it, \eg for \emph{XFeat general} an improvement of $158.51\%$ on MegaDepth and $484.70\%$ on KITTI for the AUC5. Furthermore, the XFeat refinement approach is not included in the summary table, as it can be only evaluated on XFeat*, but individual results can be found in \cref{tab:evaluation:mega,tab:evaluation:scannet,tab:evaluation:kitti}.\par

Overall, we observe that XRefine performs significantly better than existing methods, including Keypt2Subpx~\cite{keypt2subpx} and the match refinement method of PixSfM~\cite{pixsfm}. It can further be observed that \emph{XRefine specific} performs a bit better than \emph{XRefine general} which is expected as \emph{XRefine specific} is specifically trained for the respective detector,
and can therefore exploit learned priors, such as the magnitude of keypoint displacements.\par

\paragraph{Differences across datasets} 
While the performance gains achieved through refinement are significant on MegaDepth and ScanNet, we observe only small performance gains on KITTI for most detectors. This can be explained by the relatively simple visual odometry use case: in contrast to the more challenging MegaDepth and ScanNet datasets, KITTI visual odometry presents only minor visual appearance changes in the paired images. Hence, state-of-the-art feature extractors often deliver sufficiently accurate keypoints even without refinement.

\paragraph{Differences across detectors} It can further be observed that the effectiveness of match refinement depends on the keypoint detector. We observe significant performance gains for SuperPoint, DeDoDe, XFeat and XFeat*. Since SuperPoint and XFeat are providing keypoint positions only with pixel accuracy, their gain from match refinement can be expected. On the other hand, the performance of SIFT benefits only marginally, if at all, from match refinement, which could be explained by its elaborate Difference of Gaussian pyramid based keypoint detection approach.    

\paragraph{Runtime evaluation}
\Cref{tab:evaluation:runtime} shows the computation time of all refinement methods averaged over $10000$ image pairs with $2048$ $64$-dimensional XFeat~\cite{xfeat} features per image, evaluated on an Nvidia RTX A5000 GPU. 
While XRefine is with $3.61$ms only marginally slower than Keypt2Subpx with $3.43$ms, the feature-metric optimization approach of PixSFM is significantly slower with $70.28$ms. Additionally, PixSFM extracts feature embeddings for the entire images with S2DNet, which, if included in the measurement, results in a runtime of $1435.71$ms. The XFeat-Refinement approach, on the other hand, is with a runtime of $0.55$ms very light weighted, but also limited in its accuracy.

\begin{table}
\centering
\begin{tabular}{@{}l@{\hspace{4pt}}c@{\hspace{4pt}}c@{\hspace{4pt}}c@{\hspace{4pt}}c@{}}
\toprule
Refinement Method & Runtime (ms) \\
\midrule
XFeat-Refinement & 0.55 \\
Keypt2Subpx & 3.43 \\
XRefine (ours) &  3.61 \\
PixSfM without S2DNet & 70.28 \\
PixSfM with S2DNet & 1435.71 \\
\bottomrule
\end{tabular}
\caption{Runtime measurements on a NVIDIA RTX A5000.}
\label{tab:evaluation:runtime}
\end{table}

\paragraph{Ablation results}

\begin{table}
	\centering
	\begin{tabular}{@{}l@{\hspace{4pt}}c@{\hspace{4pt}}c@{\hspace{4pt}}c@{\hspace{4pt}}c@{}}
		\toprule
		Refinement & AUC5 & AUC10 & AUC20 & t(ms) \\
		\midrule
		& 37.95 & 52.43 & 64.83 & \\
		Small Specific - No Attn. & 41.20 & 54.96 & 66.32 & 2.46 \\
		Small Specific - Co-Sim & 45.58 & 59.14 & 69.72 & 3.54 \\
		Small Specific - Only 2nd & 46.53 & 59.70 & 69.88 & 3.34 \\
		\textbf{Small General} & 46.86 & 60.05 & 70.28 & 3.61 \\
		\textbf{Small Specific} & 47.52 & 60.53 & 70.68 & 3.59 \\
		Small Specific - Desc. Attn. & 47.55 & 60.76 & 70.86 & 4.34 \\
		Large General & 49.00 & 61.97 & 71.70 & 19.68 \\
		Large Specific & 50.05 & 62.82 & 72.32 & 19.71 \\
		\bottomrule
	\end{tabular}
	\caption{Results for variants of our model on MegaDepth1500~\cite{megadepth} with XFeat~\cite{xfeat} features and MNN matching. In bold, we highlight the two models for which results are presented in \cref{tab:evaluation:averages,tab:evaluation:mega,tab:evaluation:scannet,tab:evaluation:kitti}.}
	\label{tab:evaluation:ablation}
\end{table}

\begin{table}
	\centering
	\begin{tabular}{@{}l@{\hspace{4pt}}c@{\hspace{4pt}}c@{\hspace{4pt}}c@{\hspace{4pt}}c@{}}
		\toprule
		\#KP per image & Refinement & AUC5 & AUC10 & AUC20 \\
		\midrule
		2048 &  & 37.95 & 52.43 & 64.83 \\
		2048 & Keypt2Subpx & 40.46 & 54.63 & 66.31 \\
		2048 & XRefine general & \underline{46.86} & \underline{60.05} & \underline{70.28} \\
		2048 & XRefine specific & \textbf{47.52} & \textbf{60.53} & \textbf{70.68} \\
		\hline
		4096 &  & 40.16 & 54.00 & 65.62 \\
		4096 & Keypt2Subpx & 42.62 & 56.18 & 67.04 \\
		4096 & XRefine general & \underline{48.01} & \underline{60.90} & \underline{70.70} \\
		4096 & XRefine specific & \textbf{48.79} & \textbf{61.33} & \textbf{71.01} \\
		\hline
		8192 &  & 39.59 & 53.43 & 64.91 \\
		8192 & Keypt2Subpx & 41.36 & 54.86 & 65.91 \\
		8192 & XRefine general & \underline{47.19} & \underline{60.18} & \underline{69.98} \\
		8192 & XRefine specific & \textbf{47.97} & \textbf{60.62} & \textbf{70.38} \\
		\hline
		16384 &  & 39.58 & 53.29 & 64.78 \\
		16384 & Keypt2Subpx & 41.39 & 54.89 & 65.92 \\
		16384 & XRefine general & \underline{47.14} & \underline{60.04} & \underline{69.90} \\
		16384 & XRefine specific & \textbf{47.93} & \textbf{60.66} & \textbf{70.37} \\
		\bottomrule
	\end{tabular}
	\caption{Results for varying numbers of extracted keypoints (KPs) per image on MegaDepth1500~\cite{megadepth} with XFeat~\cite{xfeat} features and mutual nearest neighbor matching.}
	\label{tab:evaluation:number_of_kps}
\end{table}

\begin{table*}[t]
	\centering
	\begin{tabular}{l|ccc|ccc||ccc|ccc}
		\hline
		& \multicolumn{6}{c||}{\textbf{ETH3D indoor}} & \multicolumn{6}{c}{\textbf{ETH3D outdoor}} \\
		\textbf{Refinement Method} & \multicolumn{3}{c|}{Accuracy (\%)} & \multicolumn{3}{c||}{Completeness (\%)} & \multicolumn{3}{c|}{Accuracy (\%)} & \multicolumn{3}{c}{Completeness (\%)} \\
		& 1cm & 2cm & 5cm & 1cm & 2cm & 5cm & 1cm & 2cm & 5cm & 1cm & 2cm & 5cm \\
		\hline
		& 78.89 & 87.73 & 94.49 & 0.61 & 2.26 & 9.03 & 54.08 & 69.29 & 83.64 & 0.10 & 0.59 & 4.08 \\
		\hline
		XRefine general & 84.31 & 90.80 & 96.04 & 0.63 & 2.25 & 8.78 & 62.83 & 76.10 & 87.76 & 0.12 & 0.63 & 4.14 \\
		PixSfM KA & 89.09 & 93.55 & 96.96 & 0.71 & 2.43 & 9.19 & 70.55 & 82.39 & 91.46 & 0.15 & 0.76 & 4.70 \\
		\hline
	\end{tabular}
	\caption{Triangulation results of different refinement methods on ETH3D indoor and outdoor datasets. Our proposed $n$-view refinement consistently improves triangulation accuracy. PixSfM yields most accurate results for this use-case, as it performs a joint keypoint refinement across the full tracks, rather than separate pairwise refinements. Keypt2Subpx and the XFeat-Refinement approach cannot be applied for this use-case as it is limited to 2-view refinement which would yield inconsistent tracks.}
	\label{tab:eth3d_results}
\end{table*}

In \cref{tab:evaluation:ablation}, we present results for several variants of our proposed model for XFeat~\cite{xfeat} on MegaDepth1500~\cite{megadepth}. \emph{Small General} and \emph{Small Specific} represent \emph{XRefine general} and \emph{XRefine specific} as described in \cref{sec:method}.
Removing the cross-attention layer (Small Specific - No Attn.) significantly reduces performance as information is no longer exchanged between the matched keypoint regions. Replacing the score map head with descriptor cosine similarity (Small Specific - Co-Sim), as in Keypt2Subpx~\cite{keypt2subpx}, is marginally faster but less accurate and sacrifices generalizability, because this model requires per-descriptor training. Refining only the second keypoint (Small Specific - Only 2nd), as proposed in \cref{sec:method:n_images}, lowers accuracy slightly due to its restriction. Finally, incorporating an additional attention mechanism with the average descriptor (Small Specific - Desc. Attn.) yields a small accuracy gain but at a disproportionately increased runtime.

\emph{Large General} and \emph{Large Specific} are similar to \emph{Small General} and \emph{Small Specific}, but they make use of a larger architecture. In contrast to the small models, the large models reduce the embedding size to $5\times5$ instead of $3\times3$, by adding padding once more in the encoder. Also, they employ three cross attention blocks between the patch embeddings, instead of only one. We observe significantly improved pose estimation results for the large variants, but also a significantly increased runtime. These models could be used in use cases without strict runtime requirements.

\paragraph{Varying numbers of keypoints}
We investigate the effect of having varying numbers of keypoints extracted per image. Results for XFeat matches are presented in \cref{tab:evaluation:number_of_kps}. XFeat reaches best performance at $4096$ keypoints per image. With $21.49\%$ the relative improvement of the AUC5 metric from using \emph{XRefine specific} refinement at this number of keypoints per image is a bit smaller than it is at $2048$ keypoints per image with $25.22\%$, but still significant. For larger numbers of keypoints per image the performance of XFeat, with and without refinement, decreases slightly. The reduced advantage of using refinement with larger numbers of keypoints per image might be explained by a higher chance of obtaining a consistent set of accurate matches.

\subsection{Point cloud triangulation}
\label{sec:evaluation:triangulation}
To demonstrate the benefit of the proposed refinement in $n$-view 3D vision problems, we evaluate its effect on 3D point cloud triangulation. Using the ETH3D dataset~\cite{eth3d}, we follow the protocol from~\cite{pixsfm} and use $n$-view feature tracks to triangulate a sparse 3D model, given reference camera poses and intrinsics. Evaluation is based on the PixSfM~\cite{pixsfm} repository with SuperPoint and MNN matching, where we integrated our refinement, but deactivated the feature-metric bundle adjustment for all methods, to compare only the effect of keypoint refinement.\par
\cref{tab:eth3d_results} shows that our $n$-view refinement introduced in \cref{sec:method:n_images} consistently improves triangulation accuracy compared to no refinement, demonstrating the suitability of XRefine for 3D vision tasks beyond relative pose estimation. The improvement achieved by PixSfM is not reached which is expected as PixSfM is designed to jointly optimize all keypoints within a track, whereas our approach takes separate pairs of keypoints as input. While the joint optimization of PixSfM results in highest accuracy, it comes at the cost of computation time: While PixSfM scales quadratically with track length $T$, \ie with $\mathcal{O}(T^2)$, our pairwise refinement exhibits linear scaling $\mathcal{O}(T)$. Together with the generally higher computation time of PixSfM for a single image pair (see \cref{tab:evaluation:runtime}), this shows a trade-off between accuracy and runtime. While our refinement is significantly faster, most accurate $n$-view triangulation results can be obtained by the global refinement used in PixSfM.
\FloatBarrier
\section{Conclusion}
\label{sec:conclusion}
We presented a novel match refinement model that outperforms other state-of-the-art refinement methods in its impact on pose estimation performance without sacrificing computational efficiency. This is achieved through cross-attention between image patch embeddings without requiring detector-specific inputs like descriptors or score maps. It was shown that the model can be trained in a generalized manner, making it applicable to any keypoint detector without retraining. While extending the approach from two views to $n$ views yielded clear improvements in 3D point cloud triangulation, future work may enhance this further by adapting the architecture to directly accept $n$ image patches as input. This would enable globally optimal refinement and potentially lead to higher accuracy gains in multi-view applications. Overall, this work represents a step toward more accurate 3D vision, and can be readily incorporated into existing sparse keypoint-based systems.
{
    \small
    \bibliographystyle{ieeenat_fullname}
    \bibliography{main}

@String(IJCV = {Int. J. Comput. Vis.})

@String(CVPR= {IEEE Conf. Comput. Vis. Pattern Recog.})

@String(ICCV= {Int. Conf. Comput. Vis.})

@String(ECCV= {Eur. Conf. Comput. Vis.})

@String(NIPS= {Adv. Neural Inform. Process. Syst.})

@String(ICLR = {Int. Conf. Learn. Represent.})

@String(IJCAI = {IJCAI})

@String(CVPRW= {IEEE Conf. Comput. Vis. Pattern Recog. Worksh.})

@String(ThreeDV = {Int. Conf. on 3D Vision})

@String(IRE = {IEEE Trans. Instrum. Meas.})

@String(TMLR = {Trans. on Machine Learning Research})

@String(IROS = {IEEE/RSJ Int. Conf. on Intelligent Robots and Systems})

@String(IJCV  = {IJCV})

@String(CVPR  = {CVPR})

@String(ICCV  = {ICCV})

@String(ECCV  = {ECCV})

@String(NIPS  = {NeurIPS})

@String(ICLR  = {ICLR})

@String(CVPRW= {CVPRW})

@InProceedings{keypt2subpx,
author="Kim, Shinjeong
and Pollefeys, Marc
and Barath, Daniel",
editor="Leonardis, Ale{\v{s}}
and Ricci, Elisa
and Roth, Stefan
and Russakovsky, Olga
and Sattler, Torsten
and Varol, G{\"u}l",
title="Learning to Make Keypoints Sub-pixel Accurate",
booktitle=ECCV,
year="2025",
publisher="Springer Nature Switzerland",
pages="413--431",
}

@InProceedings{pixsfm,
    author    = {Lindenberger, Philipp and Sarlin, Paul-Edouard and Larsson, Viktor and Pollefeys, Marc},
    title     = {Pixel-Perfect Structure-From-Motion With Featuremetric Refinement},
    booktitle = ICCV,
    month     = {October},
    year      = {2021},
    pages     = {5987-5997}
}

@InProceedings{local_feature_refinement,
author="Dusmanu, Mihai
and Sch{\"o}nberger, Johannes L.
and Pollefeys, Marc",
editor="Vedaldi, Andrea
and Bischof, Horst
and Brox, Thomas
and Frahm, Jan-Michael",
title="Multi-view Optimization of Local Feature Geometry",
booktitle=ECCV,
year="2020",
publisher="Springer International Publishing",
address="Cham",
pages="670--686",
}

@inproceedings{lukas_kanade,
  title={An iterative image registration technique with an application to stereo vision},
  author={Lucas, Bruce D and Kanade, Takeo},
  booktitle=IJCAI,
  volume={2},
  pages={674--679},
  year={1981}
}

@INPROCEEDINGS{ic_lk,
  author={Baker, S. and Matthews, I.},
  booktitle=CVPR, 
  title={Equivalence and efficiency of image alignment algorithms}, 
  year={2001},
  volume={1},
  number={},
  pages={I-I},
}

@InProceedings{Lui_2018_CVPR,
author = {Lui, Vincent and Geeves, Jonathon and Yii, Winston and Drummond, Tom},
title = {Efficient Subpixel Refinement With Symbolic Linear Predictors},
booktitle = CVPR,
month = {June},
year = {2018}
}

@inproceedings{xfeat,
  title={{XFeat}: Accelerated Features for Lightweight Image Matching},
  author={Potje, G. and Cadar, F. and Araujo, A. and Martins, R. and Nascimento, E. R.},
  booktitle=CVPR,
  pages={2682--2691},
  year={2024}
}

@Article{sift,
author = "Lowe, D. G.",
title = {Distinctive Image Features from Scale-Invariant Keypoints},
journal = IJCV,
year = "2004",
month = "Nov",
day = "01",
volume = "60",
number = "2",
pages = "91--110",
}

@InProceedings{superpoint,
author = {DeTone, Daniel and Malisiewicz, Tomasz and Rabinovich, Andrew},
title = {{SuperPoint}: Self-Supervised Interest Point Detection and Description},
booktitle = CVPRW,
month = {June},
year = {2018}
}

@InProceedings{superglue,
author = {Sarlin, Paul-Edouard and DeTone, Daniel and Malisiewicz, Tomasz and Rabinovich, Andrew},
title = {{SuperGlue}: Learning Feature Matching With Graph Neural Networks},
booktitle = CVPR,
month = {June},
year = {2020}
}

@InProceedings{lightglue,
    author    = {Lindenberger, Philipp and Sarlin, Paul-Edouard and Pollefeys, Marc},
    title     = {{LightGlue}: Local Feature Matching at Light Speed},
    booktitle = ICCV,
    month     = {October},
    year      = {2023},
    pages     = {17627-17638}
}

@InProceedings{gluestick,
  title     = {{GlueStick}: Robust Image Matching by Sticking Points and Lines Together},
  author    = {R{\'e}mi Pautrat* and
               Iago Su{\'a}rez* and
               Yifan Yu and
               Marc Pollefeys and
               Viktor Larsson},
  booktitle = ICCV,
  year      = {2023}
}

@inproceedings{r2d2,
 author = {Revaud, Jerome and De Souza, Cesar and Humenberger, Martin and Weinzaepfel, Philippe},
 booktitle = NIPS,
 pages = {},
 title = {{R2D2}: Reliable and Repeatable Detector and Descriptor},
 volume = {32},
 year = {2019}
}

@inproceedings{disk,
 author = {Tyszkiewicz, Micha\l  and Fua, Pascal and Trulls, Eduard},
 booktitle = NIPS,
 pages = {14254--14265},
 title = {{DISK}: Learning local features with policy gradient},
 volume = {33},
 year = {2020}
}

@INPROCEEDINGS{dedode,
  author={Edstedt, Johan and Bökman, Georg and Wadenbäck, Mårten and Felsberg, Michael},
  booktitle=ThreeDV, 
  title={{DeDoDe}: Detect, Don’t Describe — Describe, Don’t Detect for Local Feature Matching}, 
  year={2024},
  volume={},
  number={},
  pages={148-157},
}

@InProceedings{dedodev2,
    author    = {Edstedt, Johan and B\"okman, Georg and Zhao, Zhenjun},
    title     = {{DeDoDe v2}: Analyzing and Improving the DeDoDe Keypoint Detector},
    booktitle = CVPRW,
    month     = {June},
    year      = {2024},
    pages     = {4245-4253}
}

@article{dinov2,
title={{DINO}v2: Learning Robust Visual Features without Supervision},
author={Maxime Oquab and Timoth{\'e}e Darcet and Th{\'e}o Moutakanni and Huy V. Vo and Marc Szafraniec and Vasil Khalidov and Pierre Fernandez and Daniel HAZIZA and Francisco Massa and Alaaeldin El-Nouby and Mido Assran and Nicolas Ballas and Wojciech Galuba and Russell Howes and Po-Yao Huang and Shang-Wen Li and Ishan Misra and Michael Rabbat and Vasu Sharma and Gabriel Synnaeve and Hu Xu and Herve Jegou and Julien Mairal and Patrick Labatut and Armand Joulin and Piotr Bojanowski},
journal=TMLR,
issn={2835-8856},
year={2024},
url={https://openreview.net/forum?id=a68SUt6zFt},
note={Featured Certification}
}

@ARTICLE{aliked,
  author={Zhao, Xiaoming and Wu, Xingming and Chen, Weihai and Chen, Peter C. Y. and Xu, Qingsong and Li, Zhengguo},
  journal=IRE, 
  title={{ALIKED}: A Lighter Keypoint and Descriptor Extraction Network via Deformable Transformation}, 
  year={2023},
  volume={72},
  number={},
  pages={1-16},
}

@article{unsuperpoint,
  title={{UnsuperPoint}: End-to-end unsupervised interest point detector and descriptor},
  author={Christiansen, Peter Hviid and Kragh, Mikkel Fly and Brodskiy, Yury and Karstoft, Henrik},
  journal={arXiv preprint arXiv:1907.04011},
  year={2019}
}

@inproceedings{kp2d,
   author = {Tang, Jiexiong and Kim, H. and Guizilini, V. and Pillai, S. and Rares, A.},
   booktitle = ICLR,
   title = {Neural Outlier Rejection For Self-Supervised Keypoint Learning},
   year = {2020}
}

@InProceedings{loftr,
    author    = {Sun, Jiaming and Shen, Zehong and Wang, Yuang and Bao, Hujun and Zhou, Xiaowei},
    title     = {{LoFTR}: Detector-Free Local Feature Matching With Transformers},
    booktitle = CVPR,
    month     = {June},
    year      = {2021},
    pages     = {8922-8931}
}

@InProceedings{roma,
    author    = {Edstedt, Johan and Sun, Qiyu and B\"okman, Georg and Wadenb\"ack, M\r{a}rten and Felsberg, Michael},
    title     = {{RoMa}: Robust Dense Feature Matching},
    booktitle = CVPR,
    month     = {June},
    year      = {2024},
    pages     = {19790-19800}
}

@InProceedings{megadepth,
author = {Li, Zhengqi and Snavely, Noah},
title = {{MegaDepth}: Learning Single-View Depth Prediction From Internet Photos},
booktitle = CVPR,
month = {June},
year = {2018}
}

@INPROCEEDINGS{kitti,
  author={Geiger, Andreas and Lenz, Philip and Urtasun, Raquel},
  booktitle=CVPR, 
  title={Are we ready for autonomous driving? The {KITTI} vision benchmark suite}, 
  year={2012},
  volume={},
  number={},
  pages={3354-3361},
}

@InProceedings{scannet,
author = {Dai, Angela and Chang, Angel X. and Savva, Manolis and Halber, Maciej and Funkhouser, Thomas and Niessner, Matthias},
title = {{ScanNet}: Richly-Annotated 3D Reconstructions of Indoor Scenes},
booktitle = CVPR,
month = {July},
year = {2017}
}

@inproceedings{eth3d,
  author = {Thomas Sch\"ops and Johannes L. Sch\"onberger and Silvano Galliani and Torsten Sattler and Konrad Schindler and Marc Pollefeys and Andreas Geiger},
  title = {A Multi-View Stereo Benchmark with High-Resolution Images and Multi-Camera Videos},
  booktitle = CVPR,
  year = {2017}
}

@inproceedings{dust3r,
      title={{DUSt3R}: Geometric {3D} Vision Made Easy}, 
      author={Shuzhe Wang and Vincent Leroy and Yohann Cabon and Boris Chidlovskii and Jerome Revaud},
      booktitle = CVPR,
      year = {2024}
}

@inproceedings{S2DNet,
title = {{S2DNet}: Learning Image Features for Accurate Sparse-to-Dense Matching},
author = {Hugo Germain and Guillaume Bourmaud and Vincent Lepetit},
booktitle = ECCV,
year = {2020}
}

@article{adam,
  title={Adam: A method for stochastic optimization},
  author={Kingma, Diederik P and Ba, Jimmy},
  journal={arXiv preprint arXiv:1412.6980},
  year={2014}
}

@INPROCEEDINGS{9341229,
  author={Jau, You-Yi and Zhu, Rui and Su, Hao and Chandraker, Manmohan},
  booktitle=IROS, 
  title={Deep Keypoint-Based Camera Pose Estimation with Geometric Constraints}, 
  year={2020},
  volume={},
  number={},
  pages={4950-4957},
}

@InProceedings{GC_RANSAC,
author = {Barath, Daniel and Matas, Jiří},
title = {Graph-Cut {RANSAC}},
booktitle = CVPR,
month = {June},
year = {2018}
}

@article{jin2021image,
  title={Image matching across wide baselines: From paper to practice},
  author={Jin, Yuhe and Mishkin, Dmytro and Mishchuk, Anastasiia and Matas, Jiri and Fua, Pascal and Yi, Kwang Moo and Trulls, Eduard},
  journal=IJCV,
  volume={129},
  number={2},
  pages={517--547},
  year={2021},
  publisher={Springer}
}

@inproceedings{mast3r,
  title={Grounding image matching in 3d with mast3r},
  author={Leroy, Vincent and Cabon, Yohann and Revaud, J{\'e}r{\^o}me},
  booktitle={European Conference on Computer Vision},
  pages={71--91},
  year={2024},
  organization={Springer}
}

@inproceedings{vggt,
  title={Vggt: Visual geometry grounded transformer},
  author={Wang, Jianyuan and Chen, Minghao and Karaev, Nikita and Vedaldi, Andrea and Rupprecht, Christian and Novotny, David},
  booktitle={Proceedings of the Computer Vision and Pattern Recognition Conference},
  pages={5294--5306},
  year={2025}
}
}
\clearpage
\setcounter{page}{1}
\maketitlesupplementary

\section{Results}
\paragraph{Additional feature extractors}
\Cref{tab:sup:mega,tab:sup:scannet,tab:sup:kitti} present relative pose estimation results on MegaDepth~\cite{megadepth}, ScanNet~\cite{scannet}, and KITTI~\cite{kitti} for four additional combinations of feature extractor and matcher: ALIKED~\cite{aliked} with LightGlue (LG)~\cite{lightglue} matching, DISK~\cite{disk} with LightGlue (LG) matching, DeDoDev2~\cite{dedodev2} with Double Soft Max (DSM) matching, and R2D2~\cite{r2d2} with Mutual Nearest Neighbor (MNN) matching.
Besides for ALIKED+LG on KITTI, where all refinement approaches perform very similarly, XRefine achieves the best performance. Quite striking is the performance improvement that can be achieved for ALIKED+LG on MegaDepth, where the AUC5 without refinement is $10.71\%$ and $29.41\%$ when using XRefine specific.

\begin{table}
\centering
\begin{tabular}{@{}l@{\hspace{4pt}}c@{\hspace{4pt}}c@{\hspace{4pt}}c@{\hspace{4pt}}c@{}}
\toprule
Extract+Match & Refinement & AUC5 & AUC10 & AUC20 \\
\midrule
ALIKED+LG &  & 10.71 & 21.16 & 34.02 \\
ALIKED+LG & Keypt2Subpx & 12.32 & 23.21 & 36.14 \\
ALIKED+LG & PixSfM & 23.23 & 34.34 & 45.82 \\
ALIKED+LG & XRefine general & \underline{29.31} & \underline{41.01} & \underline{52.06} \\
ALIKED+LG & XRefine specific & \textbf{29.41} & \textbf{41.07} & \textbf{52.07} \\
\hline
DISK+LG &  & 62.24 & 74.13 & 82.67 \\
DISK+LG & Keypt2Subpx & 63.77 & 75.35 & 83.53 \\
DISK+LG & PixSfM & 63.52 & 75.14 & 83.36 \\
DISK+LG & XRefine general & \underline{64.13} & \underline{75.58} & \underline{83.67} \\
DISK+LG & XRefine specific & \textbf{64.93} & \textbf{76.27} & \textbf{84.20} \\
\hline
DeDoDe2+DSM &  & 35.90 & 50.22 & 62.76 \\
DeDoDe2+DSM & Keypt2Subpx & 46.58 & 59.91 & 70.66 \\
DeDoDe2+DSM & PixSfM & 48.45 & 61.16 & 71.28 \\
DeDoDe2+DSM & XRefine general & \textbf{51.28} & \textbf{63.87} & \textbf{73.63} \\
DeDoDe2+DSM & XRefine specific & \underline{50.94} & \underline{63.58} & \underline{73.42} \\
\hline
R2D2+MNN &  & 45.13 & 57.83 & 68.02 \\
R2D2+MNN & Keypt2Subpx & 46.03 & 58.61 & 68.61 \\
R2D2+MNN & PixSfM & 46.63 & 58.77 & 68.39 \\
R2D2+MNN & XRefine general & \underline{47.88} & \underline{60.01} & \underline{69.54} \\
R2D2+MNN & XRefine specific & \textbf{49.38} & \textbf{61.45} & \textbf{70.79} \\
\bottomrule
\end{tabular}
\caption{Pose estimation result on MegaDepth~\cite{megadepth}. Bold indicates best performance and underscores second best per feature.}
\label{tab:sup:mega}
\end{table}

\begin{table}
\centering
\begin{tabular}{@{}l@{\hspace{4pt}}c@{\hspace{4pt}}c@{\hspace{4pt}}c@{\hspace{4pt}}c@{}}
\toprule
Extract+Match & Refinement & AUC5 & AUC10 & AUC20 \\
\midrule
ALIKED+LG &  & 20.73 & 38.26 & 54.84 \\
ALIKED+LG & Keypt2Subpx & 20.93 & 38.31 & 54.81 \\
ALIKED+LG & PixSfM & 20.87 & 38.27 & 54.60 \\
ALIKED+LG & XRefine general & \underline{21.70} & \underline{39.37} & \underline{55.74} \\
ALIKED+LG & XRefine specific & \textbf{21.82} & \textbf{39.50} & \textbf{55.79} \\
\hline
DISK+LG &  & 18.48 & 34.24 & 49.12 \\
DISK+LG & Keypt2Subpx & 18.77 & 34.63 & 49.63 \\
DISK+LG & PixSfM & 18.50 & 33.98 & 48.98 \\
DISK+LG & XRefine general & \underline{19.10} & \textbf{35.23} & \textbf{50.29} \\
DISK+LG & XRefine specific & \textbf{19.39} & \underline{35.21} & \underline{50.17} \\
\hline
DeDoDe2+DSM &  & 14.48 & 28.91 & 43.80 \\
DeDoDe2+DSM & Keypt2Subpx & 16.43 & 31.64 & 46.84 \\
DeDoDe2+DSM & PixSfM & 16.10 & 30.93 & 45.97 \\
DeDoDe2+DSM & XRefine general & \textbf{17.21} & \underline{32.40} & \underline{47.61} \\
DeDoDe2+DSM & XRefine specific & \underline{17.06} & \textbf{32.47} & \textbf{47.70} \\
\hline
R2D2+MNN &  & 12.14 & 24.68 & 38.56 \\
R2D2+MNN & Keypt2Subpx & \underline{12.60} & 25.23 & 38.98 \\
R2D2+MNN & PixSfM & 11.84 & 24.00 & 37.59 \\
R2D2+MNN & XRefine general & \underline{12.60} & \underline{25.59} & \underline{39.46} \\
R2D2+MNN & XRefine specific & \textbf{12.63} & \textbf{25.72} & \textbf{39.57} \\
\bottomrule
\end{tabular}
\caption{Pose estimation result on ScanNet~\cite{scannet}. Bold indicates best performance and underscores second best per feature.}
\label{tab:sup:scannet}
\end{table}

\begin{table}
\centering
\begin{tabular}{@{}l@{\hspace{4pt}}c@{\hspace{4pt}}c@{\hspace{4pt}}c@{\hspace{4pt}}c@{}}
\toprule
Extract+Match & Refinement & AUC5 & AUC10 & AUC20 \\
\midrule
ALIKED+LG &  & 82.14 & 91.07 & 95.54 \\
ALIKED+LG & Keypt2Subpx & \textbf{84.73} & \textbf{91.70} & \underline{95.63} \\
ALIKED+LG & PixSfM & 84.34 & 91.52 & 95.54 \\
ALIKED+LG & XRefine general & \underline{84.70} & \textbf{91.70} & \textbf{95.64} \\
ALIKED+LG & XRefine specific & 84.67 & 91.69 & \underline{95.63} \\
\hline
DISK+LG &  & 84.14 & 91.47 & \textbf{95.58} \\
DISK+LG & Keypt2Subpx & 84.48 & 91.52 & 95.42 \\
DISK+LG & PixSfM & 84.19 & 91.38 & 95.35 \\
DISK+LG & XRefine general & \underline{84.55} & \underline{91.57} & 95.43 \\
DISK+LG & XRefine specific & \textbf{84.63} & \textbf{91.63} & \underline{95.48} \\
\hline
DeDoDe2+DSM &  & 83.59 & 91.22 & 95.45 \\
DeDoDe2+DSM & Keypt2Subpx & 84.16 & 91.49 & 95.56 \\
DeDoDe2+DSM & PixSfM & 84.15 & 91.46 & 95.57 \\
DeDoDe2+DSM & XRefine general & \underline{84.44} & \underline{91.67} & \underline{95.67} \\
DeDoDe2+DSM & XRefine specific & \textbf{84.52} & \textbf{91.70} & \textbf{95.69} \\
\hline
R2D2+MNN &  & 83.37 & 90.99 & 95.25 \\
R2D2+MNN & Keypt2Subpx & 83.49 & 91.12 & 95.33 \\
R2D2+MNN & PixSfM & 84.28 & 91.51 & 95.60 \\
R2D2+MNN & XRefine general & \underline{84.38} & \underline{91.57} & \underline{95.64} \\
R2D2+MNN & XRefine specific & \textbf{84.56} & \textbf{91.66} & \textbf{95.68} \\
\bottomrule
\end{tabular}
\caption{Pose estimation result on KITTI~\cite{kitti} odometry. Bold indicates best performance and underscores second best per feature.}
\label{tab:sup:kitti}
\end{table}

\paragraph{Comparison of Keypt2Subpx weights}
\Cref{tab:sup:kp2subpx} presents the results that we obtained for Keypt2Subpx when using the original weights provided by the authors (\url{https://github.com/KimSinjeong/keypt2subpx/tree/master/pretrained}) and the weights that we obtained using the same training procedure as for XRefine. At this moment in time, from the combinations of feature extractors and matchers that are considered by us, only weights for SuperPoint with LightGlue matching, DeDoDe with Double Soft Max matching, XFeat with Mutual Nearest Neighbor matching, and ALIKED with LightGlue matching are available.
We observe very similar results for Keypt2Subpx with the original weights and with our weights. Only in one case (XFeat+MNN on MegaDepth) the AUC5 reached with our weights is more than $0.1$ percentage points lower than with the original weights, while in $8$ out of the $12$ evaluations our weights have slightly higher AUC5 than the original weights.
The small differences in performance can be explained by the stochastic nature of the training process.

\begin{table*}
\centering
\begin{tabular}{@{}l@{\hspace{4pt}}c@{\hspace{4pt}}c@{\hspace{4pt}}c@{\hspace{4pt}}c@{\hspace{4pt}}c@{}}
\toprule
Dataset & Extract+Match & Refinement & AUC5 & AUC10 & AUC20 \\
\midrule
MegaDepth & SP+LG &  & 58.48 & 71.41 & 80.83 \\
MegaDepth & SP+LG & Keypt2Subpx (original weights) & 60.06 & 72.70 & 81.78 \\
MegaDepth & SP+LG & Keypt2Subpx (our weights) & 60.16 & 72.73 & 81.78 \\
\hline
MegaDepth & DeDoDe+DSM &  & 34.88 & 48.64 & 60.84 \\
MegaDepth & DeDoDe+DSM & Keypt2Subpx (original weights) & 44.02 & 57.27 & 68.14 \\
MegaDepth & DeDoDe+DSM & Keypt2Subpx (our weights) & 44.86 & 58.08 & 68.84 \\
\hline
MegaDepth & XFeat+MNN &  & 36.45 & 47.89 & 57.81 \\
MegaDepth & XFeat+MNN & Keypt2Subpx (original weights) & 38.34 & 49.56 & 59.15 \\
MegaDepth & XFeat+MNN & Keypt2Subpx (our weights) & 38.01 & 49.06 & 58.52 \\
\hline
MegaDepth & ALIKED+LG &  & 10.71 & 21.16 & 34.02 \\
MegaDepth & ALIKED+LG & Keypt2Subpx (original weights) & 11.37 & 22.01 & 34.87 \\
MegaDepth & ALIKED+LG & Keypt2Subpx (our weights) & 12.32 & 23.21 & 36.14 \\
\hline
\hline
ScanNet1500 & SP+LG &  & 19.48 & 37.40 & 54.79 \\
ScanNet1500 & SP+LG & Keypt2Subpx (original weights) & 20.11 & 38.14 & 55.38 \\
ScanNet1500 & SP+LG & Keypt2Subpx (our weights) & 20.31 & 38.21 & 55.43 \\
\hline
ScanNet1500 & DeDoDe+DSM &  & 10.13 & 21.04 & 32.42 \\
ScanNet1500 & DeDoDe+DSM & Keypt2Subpx (original weights) & 11.11 & 22.71 & 34.92 \\
ScanNet1500 & DeDoDe+DSM & Keypt2Subpx (our weights) & 11.20 & 23.02 & 34.99 \\
\hline
ScanNet1500 & XFeat+MNN &  & 10.28 & 22.04 & 35.77 \\
ScanNet1500 & XFeat+MNN & Keypt2Subpx (original weights) & 11.33 & 23.58 & 37.39 \\
ScanNet1500 & XFeat+MNN & Keypt2Subpx (our weights) & 11.27 & 23.32 & 37.08 \\
\hline
ScanNet1500 & ALIKED+LG &  & 20.73 & 38.26 & 54.84 \\
ScanNet1500 & ALIKED+LG & Keypt2Subpx (original weights) & 21.01 & 38.54 & 55.04 \\
ScanNet1500 & ALIKED+LG & Keypt2Subpx (our weights) & 20.93 & 38.31 & 54.81 \\
\hline
\hline
KITTI & SP+LG &  & 83.37 & 90.84 & 95.12 \\
KITTI & SP+LG & Keypt2Subpx (original weights) & 83.59 & 90.90 & 95.12 \\
KITTI & SP+LG & Keypt2Subpx (our weights) & 83.63 & 90.93 & 95.14 \\
\hline
KITTI & DeDoDe+DSM &  & 83.98 & 91.31 & 95.42 \\
KITTI & DeDoDe+DSM & Keypt2Subpx (original weights) & 84.16 & 91.40 & 95.49 \\
KITTI & DeDoDe+DSM & Keypt2Subpx (our weights) & 84.21 & 91.42 & 95.50 \\
\hline
KITTI & XFeat+MNN &  & 81.55 & 89.99 & 94.79 \\
KITTI & XFeat+MNN & Keypt2Subpx (original weights) & 82.21 & 90.38 & 94.97 \\
KITTI & XFeat+MNN & Keypt2Subpx (our weights) & 82.42 & 90.47 & 95.00 \\
\hline
KITTI & ALIKED+LG &  & 82.14 & 91.07 & 95.54 \\
KITTI & ALIKED+LG & Keypt2Subpx (original weights) & 84.77 & 91.73 & 95.64 \\
KITTI & ALIKED+LG & Keypt2Subpx (our weights) & 84.73 & 91.70 & 95.63 \\
\bottomrule
\end{tabular}
\caption{Comparison of Keypt2Subpx results using the original weights from the authors and our retrained weights.}
\label{tab:sup:kp2subpx}
\end{table*}

\paragraph{Varying numbers of keypoints for DeDoDe}
Similarly to \cref{tab:evaluation:number_of_kps}, the \cref{tab:sup:dedode_num_kps} presents the pose estimation performance for varying numbers of extracted keypoints per image, but for DeDoDe~\cite{dedode} features.
In our evaluation, the best performance without refinement is reached at $16384$ keypoints, while the best performance with refinement is reached at $8192$ keypoints.
We observe diminishing performance gains from refinement as the number of keypoints increases. This is expected because, when fewer matches are available, the accuracy of individual keypoint correspondences plays a more critical role in pose estimation.
At $32768$ keypoints the pose estimation accuracy when using Keypt2Subpx even becomes slightly worse than without refinement.
With XRefine, on the other hand, accuracy is still increased, \eg the AUC5 with XRefine specific is about $7\%$ higher than without refinement.

\begin{table*}
\centering
\begin{tabular}{@{}l@{\hspace{4pt}}c@{\hspace{4pt}}c@{\hspace{4pt}}c@{\hspace{4pt}}c@{}}
	\toprule
	\#KP per image & Refinement & AUC5 & AUC10 & AUC20 \\
	\midrule
	2048 &  & 38.54 & 53.92 & 67.75 \\
	2048 & Keypt2Subpx & 47.75 & 62.87 & 74.68 \\
	2048 & XRefine specific & \textbf{52.34} & \textbf{66.65} & \textbf{77.56} \\
	2048 & XRefine general & \underline{52.32} & \underline{66.38} & \underline{77.04} \\
	\hline
	4096 &  & 45.05 & 59.90 & 72.44 \\
	4096 & Keypt2Subpx & 51.65 & 65.94 & 77.21 \\
	4096 & XRefine specific & \textbf{55.67} & \textbf{69.10} & \textbf{79.18} \\
	4096 & XRefine general & \underline{55.20} & \underline{68.73} & \underline{78.79} \\
	\hline
	8192 &  & 49.63 & 63.57 & 74.75 \\
	8192 & Keypt2Subpx & 52.82 & 66.62 & 77.25 \\
	8192 & XRefine specific & \textbf{56.51} & \textbf{69.80} & \textbf{79.33} \\
	8192 & XRefine general & \textbf{56.51} & \underline{69.29} & \underline{78.84} \\
	\hline
	16384 &  & 51.68 & 65.03 & 75.49 \\
	16384 & Keypt2Subpx & 52.66 & 65.90 & 76.10 \\
	16384 & XRefine specific & \underline{56.29} & \textbf{69.06} & \textbf{78.29} \\
	16384 & XRefine general & \textbf{56.32} & \underline{68.69} & \underline{77.99} \\
	\hline
	32768 &  & 51.26 & 63.95 & 73.53 \\
	32768 & Keypt2Subpx & 50.49 & 63.62 & 73.70 \\
	32768 & XRefine specific & \textbf{54.95} & \textbf{67.13} & \textbf{76.08} \\
	32768 & XRefine general & \underline{54.69} & \underline{66.91} & \underline{75.81} \\
	\bottomrule
\end{tabular}
\caption{Results for varying numbers of extracted keypoints (KPs) per image on MegaDepth1500~\cite{megadepth} with DeDoDe~\cite{dedode} features and double soft max matching.}
\label{tab:sup:dedode_num_kps}
\end{table*}

\begin{figure*}[t]
\centering
\includegraphics[width=0.78\textwidth]{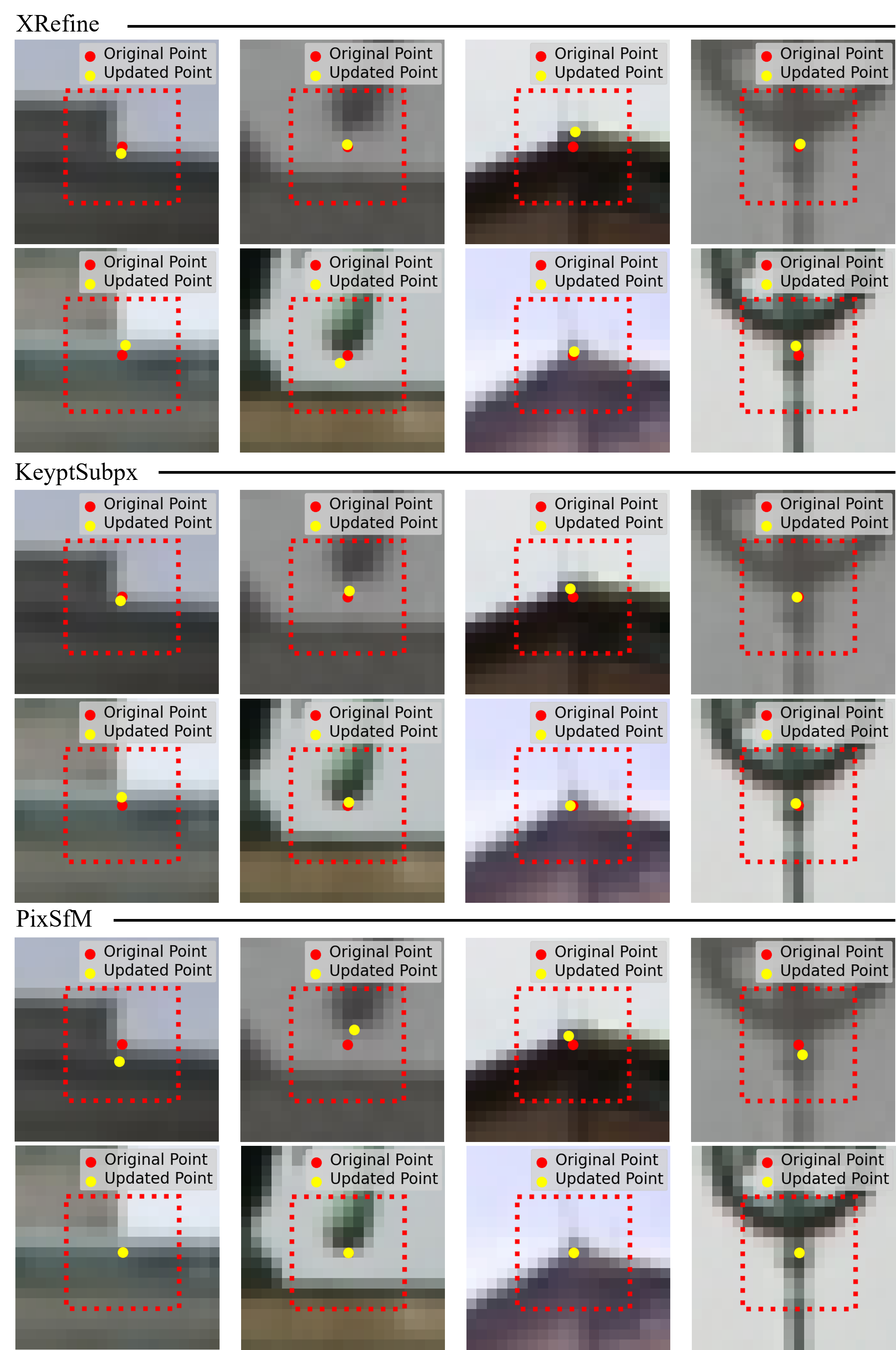}
\caption{Example keypoint refinements for XRefine (top two rows), Keypt2Subpx (middle two rows), and PixSfM (bottom two rows). Keypoints are extracted from MegaDepth, using SuperPoint and LightGlue. Each column represents the extracted patches for a given pair of matched keypoints. The same four extracted keypoint matches are refined by the three refinement methods. The presented patches have a size of $21\times21$ pixel, while the $11\times11$ area that is given as input to XRefine and Keypt2Subpx is highlighted by the red dotted square.}
\label{fig:samples}
\end{figure*}

\paragraph{Qualitative results}
\Cref{fig:samples} presents visualizations of four refinement examples for our XRefine, Keypt2Subpx~\cite{keypt2subpx}, and PixSfM~\cite{pixsfm}. 

\end{document}